\documentclass{Interspeech2024}

\interspeechcameraready

\title{Is One Brick Enough to Break the Wall of Spoken Dialogue State Tracking?}

\name[affiliation={1,2}]{Lucas}{Druart}
\name[affiliation={1}]{Valentin}{Vielzeuf}
\name[affiliation={2}]{Yannick}{Estève}

\address{
  $^1$Orange Innovation, France\\
  $^2$LIA - Avignon Université, France}
\email{lucas1.druart@orange.com, first.last@univ-avignon.fr}


\usepackage{nth}
\usepackage{subcaption}

\begin{document}

\maketitle

\begin{abstract}
    
    In Task-Oriented Dialogue (TOD) systems, correctly updating the system's understanding of the user's requests (\textit{a.k.a} dialogue state tracking) is key to a smooth interaction. Traditionally, TOD systems perform this update in three steps: transcription of the user's utterance, semantic extraction of the key concepts, and contextualization with the previously identified concepts. Such cascade approaches suffer from cascading errors and separate optimization. End-to-End approaches have been proven helpful up to the turn-level semantic extraction step. This paper goes one step further and provides (1) a novel approach for completely neural spoken DST, (2) an in depth comparison with a state of the art cascade approach and (3) avenues towards better context propagation. Our study highlights that jointly-optimized approaches are also competitive for contextually dependant tasks, such as Dialogue State Tracking (DST), especially in audio native settings. Context propagation in DST systems could benefit from training procedures accounting for the previous' context inherent uncertainty.
\end{abstract}

\section{Introduction}

    Digitization enables many tasks to be automated, nevertheless users sometimes require assistance to perform complex tasks such as making a reservation at a restaurant or booking a hotel room. Task-Oriented Dialogue (TOD) systems are designed to assist such users. A common approach to implement them is to break the problem down to three iterative steps \cite{TurDeMori_SDS}: updating the system's understanding of the users' requests, reasoning over a database and domain knowledge to choose the next action and providing the user an answer. This paper focuses on the understanding step.

\begin{figure}[htb!]
        \centering
        \includegraphics[width=\linewidth]{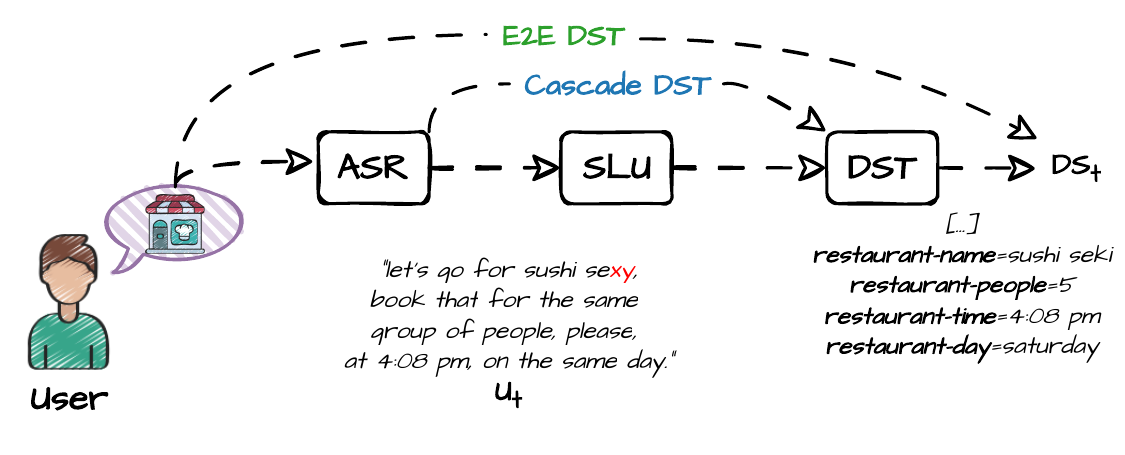}
        \caption{Spoken Dialogue State Tracking alternatives. Red characters indicate potential cascading errors.}
        \label{fig:sds}
    \end{figure}

    Traditionally the user's requests update consists of three components, respectively performing the transcription of the user's utterance, semantic extraction and contextualization of the extracted concepts \cite{williamsDialogStateTracking2016}. Unfortunately, this method presents the inconvenience of propagating errors of a component on to the next one(s) (\textit{i.e.} cascading errors) and of not optimizing all components on the final objective (\textit{i.e.} separate optimization) \cite{mdhaffar2022ImpactAnalysis}, as illustrated in Figure \ref{fig:sds}. End-to-End (E2E) approaches may address these issues by designing models in which the gradient (\textit{i.e.} error signal) can back-propagate from the output all the way to the input \cite{Serdyuk_E2ESLU_2018}. 

    On the one hand, with the advent of deep-learning and textual embeddings, state of the art Dialogue State Tracking (DST) models now work directly on automatic transcriptions \cite{jacqmin2023firstDSTC11}. However, such approaches require careful, dataset specific, mechanisms to catch and correct transcription errors together with data augmentation to increase the downstream model's robustness to specific upstream errors \cite{faruqui-hakkani-tur-2022-revisiting}.

    On the other hand, Spoken Language Understanding (SLU) directly from the speech signal, has successfully been applied to tasks which process utterances individually such as voice command slot filling \cite{STOP_E2E_SLU_Watanabe} and dialogue act classification \cite{DialogueHistory4DialogueAct, ganhotra21_interspeech_DialogueHistoryDA}.
    Such systems often leverage transfer learning of previously trained models. This is challenging because it requires a trade-off between learning new knowledge (\textit{e.g.} domain's vocabulary, domain's ontology structure) and keeping previous capabilities (\textit{e.g.} transcription of open vocabulary concepts) on a small amount of data \cite{mdhaffar2022ImpactAnalysis}.  
    
    In TOD systems the semantic extraction also depends on the dialogue's current context. While E2E SLU has been efficiently designed for single independent utterance processing \cite{STOP_E2E_SLU_Watanabe,SLURP_E2E_SLU,E2ESLU_MEDIA}, contextually dependant E2E SLU remains unexplored to the best of our knowledge. Indeed, dialogue history integration to guide the current turn's prediction (\textit{e.g.} better spelling of technical vocabulary) has already been implemented \cite{DialogueHistory4DialogueAct, ganhotra21_interspeech_DialogueHistoryDA,MEDIADialogueHistoryIntegration,sunder23_interspeech}. Yet, the tasks described in these studies can be achieved without resorting to the context which is impossible for DST (\textit{e.g.} processing cross-turn reference resolution).

    Producing high quality annotated dialogue datasets is expensive because of the cognitive load required to analyse the context and adapt the annotation. Such datasets exist for chat based dialogue understanding \cite{dmr2022, Budzianowski2018MultiWOZA} but lack for spoken dialogues, explaining the gap between E2E SLU and DST. Recently, two datasets have been introduced in an attempt to fill this gap: Spoken MultiWOZ \cite{soltau-etal-2023-dstc} and SpokenWOZ \cite{si2023spokenwoz}.
    
    This paper lies at the intersection of both directions. We focus on contextually dependant semantic extraction, such as DST, in which the previous dialogue context is mandatory to correctly process the current one (\textit{e.g.} cross-turn references resolution).
    In fact, the spoken DST models presented in this paper output a summary of the user's requests since the beginning of the dialogue. State of the art DST systems use cascade approaches which create a textual bottleneck both in terms of data and model inference. E2E approaches do not require ground-truth transcriptions and can be jointly optimized. 
    
    This paper paves the path towards E2E spoken DST with: (1) a novel completely neural DST approach, (2) a detailed comparison with a state-of-the-art cascade approach and (3) avenues towards better context propagation.
    

\section{Method}

    \subsection{Task-Oriented Dialogues}

    In TODs users require assistance from an agent to complete a task such as making a reservation at a restaurant or booking a hotel room. More formally, let us define a TOD as a sequence of $t$ dialogue turns $U_1, A_2, \dots,A_{t-1}, U_t$ where $A_{t-1}$ and $U_t$ respectively correspond to agent's turn $t-1$ and user's turn $t$. The goal of DST is to keep up to date a condensed representation of the user's requests. In this paper, users requests are represented as Dialogue States (DS) and correspond to a list of $n$ slot-value pairs flattened as \texttt{slot$_1$=value$_1$;...;slot$_n$=value$_n$}. At a given turn $t$, a DST system is thus inputted the previous context $DS_{t-2}$ and both agent's and user's most recent turns $A_{t-1}$ and $U_t$ from which it should output the updated user's requests $DS_t$\footnote{For $t = 0$, both the context and agent turns are empty.}.

    \subsection{Context Propagation}

\begin{figure}[ht!]
        \centering
        \includegraphics[width=\linewidth]{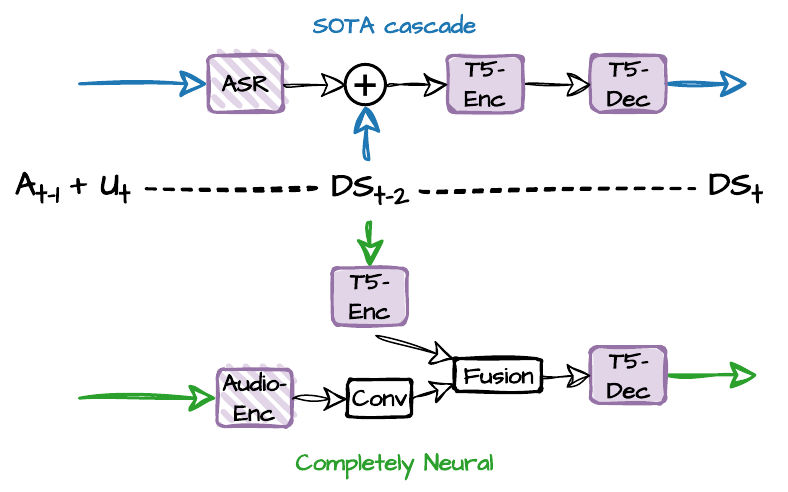}
        \caption{Two approaches for context propagation in spoken DST: SOTA cascade (top) and completely neural models (bottom). The inputs are displayed in the middle: agent previous turn $A_{t-1}$, user current turn $U_{t}$ and previous dialogue state $DS_{t-2}$. The output is the current dialogue state $DS_t$. Hatched components are speech-related while solid ones are text-related. Colored blocks are fine-tuned while white ones are trained from scratch.}
        \label{fig:e2edst}
    \end{figure}

    As the dialogue unfolds, the user might refer to previously mentioned entities. In order to design a contextually dependant SLU model, we need to propagate the context of the previous turns $DS_{t-2}$ to inform the prediction $\hat{DS}_t$ associated to the current user turn $U_t$. This paper compares two alternatives of spoken DST models detailed in the following sections, and illustrated in Figure \ref{fig:e2edst}. For a fairer comparison, we use the same pre-trained components for both approaches and train the model(s) with the same consumer grade 24Gb GPU\footnote{Code will be made available upon publication.}.

    \subsubsection{Traditional Cascade Approach}

    The cascade approach consists of an Automatic Speech Recognition (ASR) model which transcribes the agent and user's turns, respectively $A_{t-1}$ and $U_t$, concatenates them to the previous' turns context $DS_{t-2}$, and uses it as input of a Natural Language Understanding (NLU) model which predicts the updated Dialogue State $\hat{DS_t}$. 

    In our experiments, we consider two ASR models: a WavLM model \cite{Chen2021WavLMLS} fine-tuned (with two additional linear layers outputting tokens' probabilities and CTC loss) on the dataset's transcriptions and an off-the-shelf Whisper \cite{Radford2022Whisper} model. Regarding the NLU component, we focus on a T5 Encoder-Decoder \cite{RaffelT5} model. Note that the NLU model is trained with user turns transcriptions of the ASR model in order to be as close as possible to its inference regime.

    \subsubsection{Completely Neural Approach}
    
    The completely neural approach leverages the same pre-trained components. It removes the textual bottleneck by fusing the current dialogue turns with the context later on, in an embedding high dimensional space. This enriched context is then used to condition T5's decoder generation. The goal of this approach is to enable joint optimization of all components.

    An audio encoder $E_{audio}$ (\textit{e.g.} WavLM or Whisper's encoder) and a textual encoder $E_{text}$ (\textit{e.g.} T5's encoder) respectively encode the current agent and user dialogue turns (audio) and the dialogue's context in the form of the previous dialogue state $DS_{t-2}$ (textual). Given that both models do not have the same processing windows, two convolution layers (stride $3$, kernel size $9$) are added to down-sample the audio encoder's outputs. The fusion layer is a self-attention layer over the concatenation, noted $||$, of both encoder's outputs. The goal is to enable the model to select and mix the information from both encoders. Finally, a textual decoder (\textit{e.g.} T5's decoder) predicts $\hat{DS_t}$ conditioned on the fusion of both encoders' outputs.
    More formally, we have:
    \begin{align*}
        h_{state} &= E_{text} (DS_{t-2}) \\
        h_{turns} &= \text{Conv}(E_{audio} (A_{t-1} + U_t)) \\
        h &= \text{Self-Attention}(h_{state} || h_{turns}) \\
        \hat{DS}_t &= w_1 \dots w_n \\ 
        \text{ with } w_i &= argmax_{w} \ p(w | w_{i-1} \dots, w_1, h)
    \end{align*}
    
\section{Results}

    \subsection{Datasets}
    
    MultiWOZ is a human-human chat-based English Task-Oriented Dialogue (TOD) dataset commonly used for training and evaluating dialogue systems \cite{Budzianowski2018MultiWOZA}. A spoken version with vocalized user turns was published in the context of the \textit{Speech Aware Dialogue Systems} track of the \nth{11} edition of the Dialogue System Technology Challenge\footnote{https://dstc11.dstc.community/} (DSTC11) \cite{soltau-etal-2023-dstc}. Given that only the user turns are vocalized, the agent turns are concatenated as context with the previous dialogue state in our models.

    \begin{table*}[htb!]
        \begin{subtable}[h]{0.48\linewidth}
            \centering
            \resizebox{\textwidth}{!}{\begin{tabular}{lcc|cc}
                 & \multicolumn{2}{c}{\textbf{Dev}} & \multicolumn{2}{c}{\textbf{Test}} \\ \midrule
                 \textbf{Cascade (g.t. text)} & \multicolumn{2}{c}{71.4} & \multicolumn{2}{c}{71.2} \\
                 & \multicolumn{2}{c}{\textit{[70.3, 72.6]}} & \multicolumn{2}{c}{\textit{[70.1, 72.4]}} \\
                 & \textbf{TTS} & \textbf{Human} & \textbf{TTS} & \textbf{Human} \\ \midrule
                 \textbf{dstc11 baseline \cite{soltau-etal-2023-dstc}} & 38.4 & 31.8 & \multicolumn{2}{c}{n/a} \\ 
                 \textbf{dstc11 best \cite{jacqmin2023firstDSTC11}} & 47.2 & 43.2 & 44.0 & 39.5 \\ \midrule
                 \textbf{Cascade (WavLM)} & 58.2 & 55.0 & 57.2 & 53.5 \\ 
                 & \textit{[57.2, 59.3]} & \textit{[53.9, 56.2]} & \textit{[56.0, 58.3]} & \textit{[52.3, 54.7]} \\
                 \textbf{Cascade (Whisper)} & \textbf{63.7} & \textbf{63.6} & \textbf{64.4} & \textbf{62.3} \\ 
                 & \textit{[62.5, 64.8]} & \textit{[62.4, 64.8]} & \textit{[63.3, 65.6]} & \textit{[61.1, 63.5]} \\
                 \textbf{E2E (WavLM)} & 56.4 & 54.0 & 53.4 & 53.0 \\
                 & \textit{[55.3, 57.4]} & \textit{[52.9, 55.1]} & \textit{[52.3, 54.5]} & \textit{[51.8, 54.2]} \\
                 \textbf{E2E (Whisper)} & 59.0 & 56.9 & 58.3 & 56.6 \\
                 & \textit{[58.0, 60.2]} & \textit{[55.7, 58.0]} & \textit{[57.2, 59.4]} & \textit{[55.5, 57.7]} \\ \bottomrule
            \end{tabular}}
            \caption{Spoken MultiWOZ: ground-truth previous state $DS_{t-2}$}
            \label{tab:oracleDST_multiwoz_gt}
        \end{subtable}
        \hspace{\fill}
        \begin{subtable}[h]{0.48\linewidth}
            \centering 
            \resizebox{\textwidth}{!}{\begin{tabular}{lcc|cc}
                 & \multicolumn{2}{c}{\textbf{Dev}} & \multicolumn{2}{c}{\textbf{Test}} \\ \midrule
                 \textbf{Cascade (g.t. text)} & \multicolumn{2}{c}{32.0} & \multicolumn{2}{c}{30.3} \\
                 & \multicolumn{2}{c}{\textit{[30.3, 33.7]}} & \multicolumn{2}{c}{\textit{[28.5, 31.9]}} \\
                 & \textbf{TTS} & \textbf{Human} & \textbf{TTS} & \textbf{Human} \\ \midrule
                 \textbf{Cascade (WavLM)} & 19.5 & 16.2 & 17.6 & 15.3 \\ 
                 & \textit{[18.4, 20.7]} & \textit{[15.1, 17.2]} & \textit{[17.2, 19.3]} & \textit{[15.2, 17.4]} \\
                 \textbf{Cascade (Whisper)} & \textbf{24.0} & \textbf{21.9} & \textbf{23.1} & \textbf{21.3} \\ 
                 & \textit{[22.7, 25.3]} & \textit{[20.6, 23.2]} & \textit{[21.9, 24.4]} & \textit{[20.0, 22.6]} \\
                 \textbf{E2E (WavLM)} & 15.1 & 14.4 & 13.7 & 14.6 \\
                 & \textit{[14.1, 16.0]} & \textit{[13.4, 15.4]} & \textit{[12.8, 14.6]} & \textit{[13.6, 15.6]} \\
                 \textbf{E2E (Whisper)} & 19.1 & 17.6 & 18.5 & 16.6 \\
                 & \textit{[18.0, 20.2]} & \textit{[16.5, 18.7]} & \textit{[17.4, 19.5]} & \textit{[15.7, 17.7]} \\ \bottomrule
            \end{tabular}}
            \caption{Spoken MultiWOZ: predicted previous state $\hat{DS}_{t-2}$}
            \label{tab:oracleDST_multiwoz_pred}
        \end{subtable}
        
        \begin{subtable}[h]{0.5\linewidth}
            \centering
            \begin{tabular}{lcc}
                 & \textbf{Dev} & \textbf{Test} \\ \midrule
                 \textbf{Cascade (WavLM)} & \textbf{82.3} & 63.0 \\ 
                 & \textit{\footnotesize[81.3, 83.3]} & \textit{\footnotesize[61.7, 64.3]} \\
                 \textbf{Cascade (Whisper)} & 80.7 & 64.2 \\ 
                 & \textit{\footnotesize[79.5, 81.8]} & \textit{\footnotesize[62.8, 65.5]} \\
                 \textbf{E2E (WavLM)} & 70.7 & 61.8 \\
                 & \textit{\footnotesize[69.4, 72.0]} & \textit{\footnotesize[60.7, 63.0]} \\
                 \textbf{E2E (Whisper)} & 81.6 & \textbf{80.5} \\ 
                 & \textit{\footnotesize[80.4, 82.8]} & \textit{\footnotesize[79.6, 81.3]} \\ \bottomrule
            \end{tabular}
            \caption{SpokenWOZ: ground-truth previous state $DS_{t-2}$}
            \label{tab:oracleDST_spokenwoz_gt}
        \end{subtable}
        \hspace{\fill}
        \begin{subtable}[h]{0.5\linewidth}
            \centering
            \begin{tabular}{lcc}
                 & \textbf{Dev} & \textbf{Test} \\ \midrule
                 \textbf{Cascade (WavLM)} & 24.6 & 23.4 \\ 
                 & \textit{\footnotesize[23.0, 26.3]} & \textit{\footnotesize[22.4, 24.6]} \\
                 \textbf{Cascade (Whisper)} & 24.3 & 23.5 \\ 
                 & \textit{\footnotesize[22.8, 25.8]} & \textit{\footnotesize[22.5, 24.6]} \\
                 \textbf{E2E (WavLM)} & 22.2 & 20.3 \\ 
                 & \textit{\footnotesize[20.7, 23.7]} & \textit{\footnotesize[19.3, 21.3]} \\
                 \textbf{E2E (Whisper)} & \textbf{26.5} & \textbf{24.1} \\ 
                 & \textit{\footnotesize[24.7, 28.5]} & \textit{\footnotesize[23.1, 25.2]} \\ \bottomrule
            \end{tabular}
            \caption{SpokenWOZ: predicted previous state $\hat{DS}_{t-2}$}
            \label{tab:oracleDST_spokenwoz_pred}
        \end{subtable}
    
    \caption{JGA$\uparrow$ with bootstrapped 95\% confidence intervals. \textbf{Cascade (g.t. text)} shows upper-bound performance on the ground-truth transcriptions. Note that \cite{soltau-etal-2023-dstc} and \cite{jacqmin2023firstDSTC11} use the complete dialogue history as input to the DST model which is not possible for audio native E2E approaches \cite{sunder23_interspeech}.}
    \label{tab:DST}
    \end{table*}
    
    The user utterances in the training set are available as synthetic speech, whereas the dev and test sets (\textbf{Dev}$|$\textbf{Test}) include both synthetic and human speech versions (\textbf{TTS}$|$\textbf{Human}). The dataset contains close to 10,000 dialogues with a 80/10/10 train-dev-test split and an average of 13.3 turns per dialogue. Among the pre-defined slots, we can distinguish 3 groups: categorical slots with a closed set of values ($\sim$60\%), non-categorical slots with an open set of values ($\sim$30\%) and time slots ($\sim$10\%). Note that, in order to reduce the value overlap across sets, non-categorical slots were replaced and time slots offset in the \textbf{Dev} and \textbf{Test} sets.

    SpokenWOZ \cite{si2023spokenwoz} is a human-human multi-domain spoken English TOD dataset. It extends the MultiWOZ's set of slots with cross-turn and reasoning slots. It contains 5,700 dialogue recordings with a 4200/500/1000 train/dev/test split which corresponds to a total of 203,074 dialogue turns and 249 hours of audio. Given the native audio nature of the dataset, no ground-truth transcriptions are available. We thus consider the dataset's provided ASR transcriptions instead of the outputs of a fine-tuned WavLM. 

    Given the low quantity of data at our disposal, we use 100\% of the training sets for both dataset.

    \subsection{Evaluation}

\begin{figure}[htb!]
        \centering
        \begin{subfigure}[h]{0.5\textwidth}
            \includegraphics[width=\linewidth]{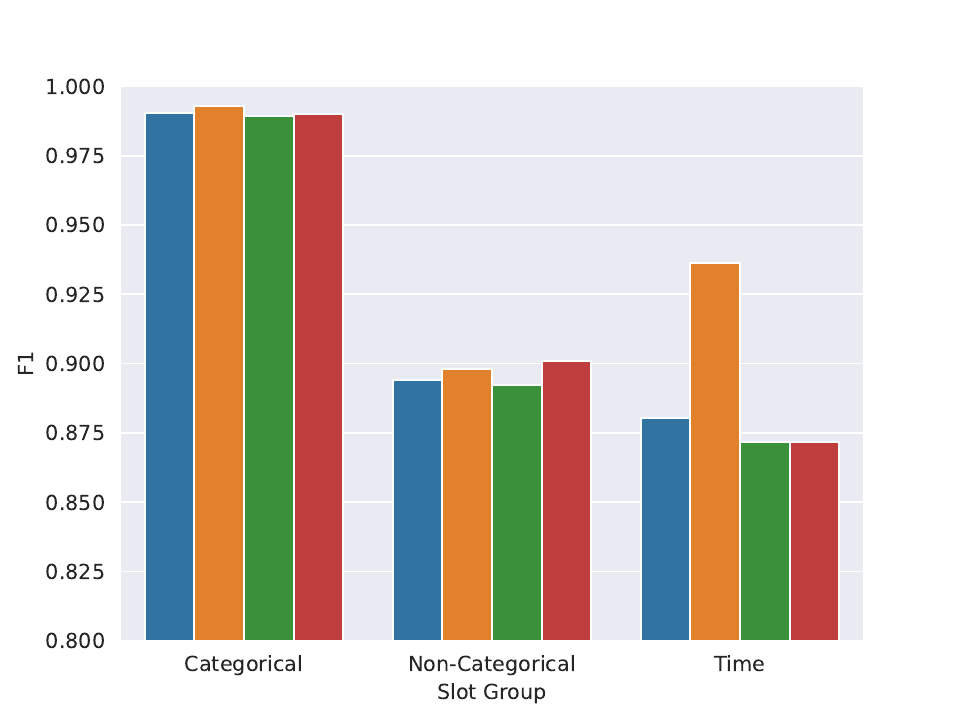}
            \caption{Spoken MultiWOZ Test-Human}
            \label{fig:f1_multiwoz}
        \end{subfigure}
        \begin{subfigure}[h]{0.5\textwidth}
            \includegraphics[width=\linewidth]{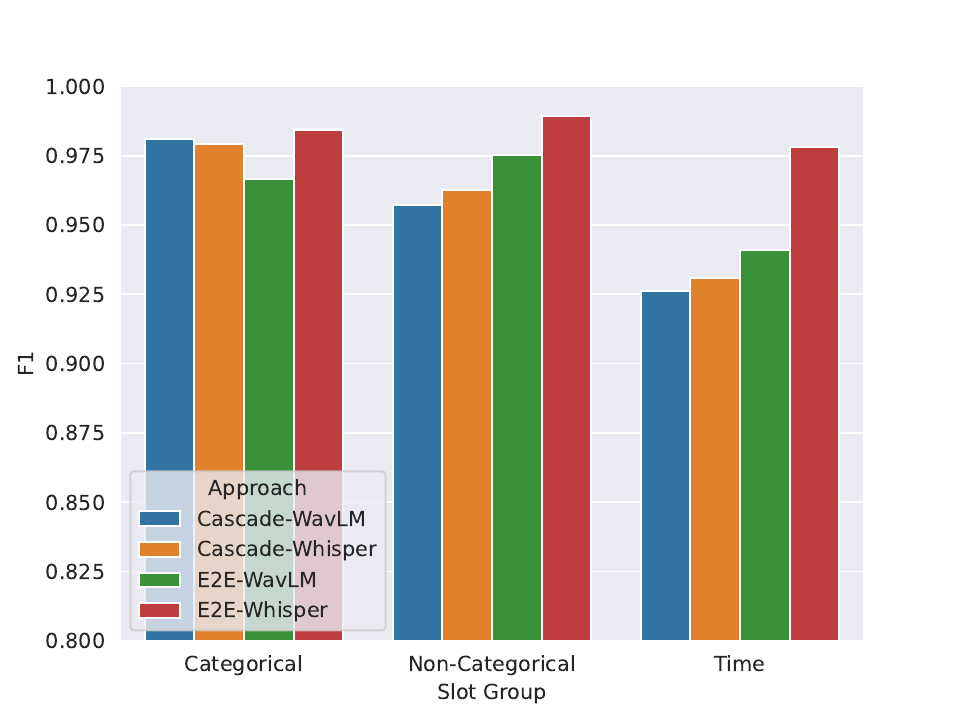}
            \caption{SpokenWOZ Test}
            \label{fig:f1_spokenwoz}
        \end{subfigure}
        \caption{Slot group average F1}
        \label{fig:f1s}
    \end{figure}

\begin{figure}[htb!]
        \centering
        \begin{subfigure}[h]{0.5\textwidth}
            \includegraphics[width=\linewidth]{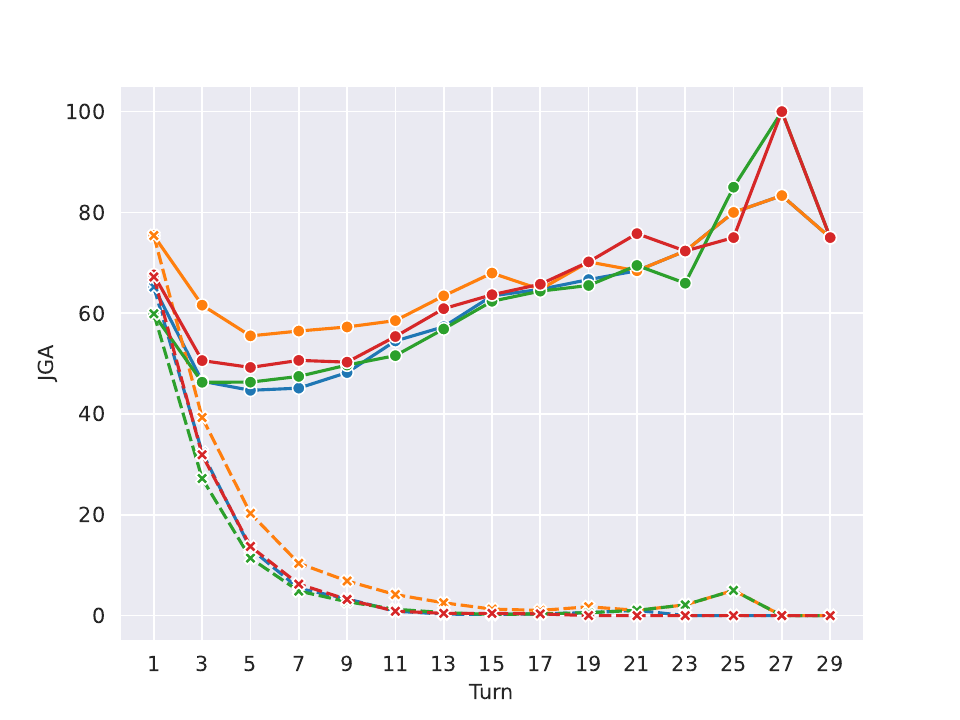}
            \caption{Spoken MultiWOZ Test-Human}
            \label{fig:turn_acc_multiwoz}
        \end{subfigure}
        \begin{subfigure}[h]{0.5\textwidth}
            \includegraphics[width=\linewidth]{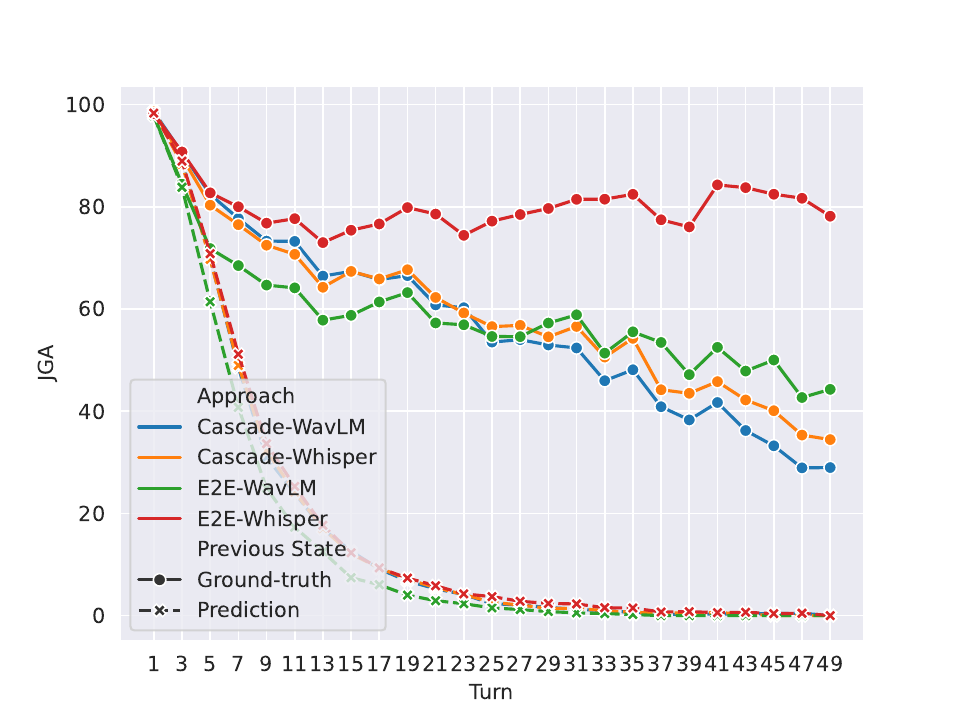}
            \caption{SpokenWOZ Test}
            \label{fig:turn_acc_spokenwoz}
        \end{subfigure}
        \caption{Turn accuracy with and without ground-truth previous state for each approach. Note that there are fewer and fewer dialogues as the number of turns increases.}
        \label{fig:turn_acc}
        \vspace{-0.6cm}
    \end{figure}

    We evaluate all approaches with a turn-level exact match metric known as Joint-Goal Accuracy (JGA$\uparrow$)~\cite{zhong-etal-2018-global}. 
    This metric requires to post-process the coma separated slot-value output format to convert it into a valid dictionary which does not take into account the order of the slot-value pairs. We present the results of all three approaches in two scenarios: with ground truth previous context $DS_{t-2}$, and with the previously predicted context $\hat{DS}_{t-2}$ in Table \ref{tab:DST}. We further analyse the performance per slot group in Figure \ref{fig:f1s} and per dialogue turn in Figure \ref{fig:turn_acc}.

    Table \ref{tab:DST} highlights that the joint optimization does indeed seem to robustify the WavLM audio encoder in the sense that it reduces the performance gap between TTS and Human for spoken MultiWOZ. Moreover, the completely neural approach leveraging a robust audio encoder such as Whisper's performs on par with cascade approaches and even slightly better in an audio native setting such as with SpokenWOZ.

    \subsubsection{Slot group analysis}

    In order to get a more precise understanding of the differences between these approaches, we further evaluate each slot's F1-measure. We present each slot group's average F1-measure on the spoken MultiWOZ \textbf{Test-Human} and SpokenWOZ \textbf{Test} sets in Figure \ref{fig:f1s}. Categorical slots present little difficulty while non-categorical and time slots are more challenging especially when an effort is made to have less overlap between those slots' values such as in spoken MultiWOZ. In such a setting, Whisper's transcription formatting seems appreciated for time slots values. In a native audio setting, such as SpokenWOZ, the end-to-end approaches perform better than cascade approaches which underlines the advantage of joint-optimization in such settings. In addition, Whisper's encoder seems to improve time formatting capabilities suggesting that part of the formatting information might be already present in its encoder.

    \subsubsection{Dialogue turn analysis}

    When comparing the per-turn performance, we observe that, as illustrated in Figure \ref{fig:turn_acc}, the robustness of the audio encoder prevents a too fast collapse of the turn accuracy as the dialogue unfolds. However, in a more realistic scenario where we base our next prediction on the previous one $\hat{DS}_{t-2}$\footnote{Note that $A_{t-1}$ and $U_t$ remain unchanged which might lead to some incoherences.}, all approaches have trouble following the course of the dialogue. This suggests that additional training mechanisms such as done for other sequence prediction tasks~\cite{bengio2015scheduled} might be required to compensate this training-inference discrepancy.  
    
    When comparing datasets we also notice that the native audio setting of SpokenWOZ underlines the advantage of joint-optimization since the completely neural approaches perform better on further dialogue turns. Note that only the completely neural approach with Whisper's encoder seems to perform roughly equally among dialogue turns. This suggests that this encoder captures better the details necessary to update the users' requests. 

    \section{Discussion}

    This paper focuses on spoken DST, for which, to the best of our knowledge, only two datasets are available. While E2E approaches often require more data to reach the same level of performance \cite{lugosch2019E2EPre-training}, this paper compares cascade and E2E approaches at the same resource level. Meaning that we use $100\%$ of each dataset's training set and the same pre-trained backbone models for both approaches. In addition, both datasets assume the dialogue turns to be perfectly separable. Future datasets will enable to study E2E approaches in higher resource and more realistic settings.
    Given the exact match nature of JGA, a more fine-grained evaluation to assess which errors are low-impact errors (\textit{e.g.} rectified with the help of a database, with no impact on the dialogue trajectory) and an adapted post processing of the non categorical slot values is left as future work.

    \section{Conclusion}

    In order to pave the path towards E2E spoken DST, we analyze the differences between a state of the art cascade approach and a completely neural approach. Our study highlights that although the cascade approach remains the most accurate approach, completely neural approaches are competitive especially in audio native settings such as SpokenWOZ. However, context propagation in completely neural approaches remains an open challenge. Integrating the previous context's uncertainty into the training process such as done for sequence prediction tasks \cite{bengio2015scheduled} seems an interesting step in this direction. 

\bibliographystyle{IEEEtran}
\bibliography{refs}

\end{document}